# Labels Are Not Perfect: Improving Probabilistic Object Detection via Label Uncertainty

Di Feng[1,2], Lars Rosenbaum[1], Fabian Timm[1], and Klaus Dietmayer[2]

[1] Corporate Research, Robert Bosch GmbH, 71272, Renningen, Germany
{Di.Feng, Lars.Rosenbaum, Fabian.Timm}@de.bosch.com
[2] Institute of Measurement, Control and Microtechnology, Ulm University, 89081, Ulm, Germany.
klaus.dietmayer@uni-ulm.de

**Abstract.** Reliable uncertainty estimation is crucial for robust object detection in autonomous driving. However, previous works on probabilistic object detection either learn predictive probability for bounding box regression in an un-supervised manner, or use simple heuristics to do uncertainty regularization. This leads to unstable training or suboptimal detection performance. In this work, we leverage our previously-proposed method for estimating uncertainty inherent in ground truth bounding box parameters (which we call label uncertainty) to improve the detection accuracy of a probabilistic LiDAR-based object detector. Experimental results on the KITTI dataset show that our method surpasses both the baseline model and the models based on simple heuristics by up to 3.6% in terms of Average Precision.

**Keywords:** Probabilistic object detection, LiDAR, autonomous driving

## 1 Introduction

Capturing reliable uncertainty in object detection networks is indispensable for safe autonomous driving [15]. In recent years, many probabilistic object detectors have been proposed [3, 4, 18, 10, 21, 16, 6]. A prevalent method is to assume a certain probability distribution over the detector outputs (e.g. the Gaussian distribution), and to learn the parameters for such a distribution by minimizing the negative log likelihood (NLL). However, this direct-modelling strategy learns predictive probability in an un-supervised manner, which may lead to unstable training or suboptimal detection performance [21].

He *et al.* [12] and Meyer *et al.* [17] tackle this problem by minimizing the Kullback-Leibler Divergence (KLD) between a simple prior probability distribution and the predictive probability. In this way, the network is regularized to predict probability close to the prior distribution. Since such prior distribution is often related to the generation process of ground truth bounding box parameters, it is also referred to as the label uncertainty. Though the effectiveness of KLD is highly dependent on the quality of label uncertainty, both methods approximate it with simple heuristics, which do not fully reflect its behaviours: He *et al.* [12] ignore label noises by setting a Dirac-delta function on labels. In this case, the derivatives of the KLD loss degenerates to those of the NLL loss. Meyer *et al.* [17] approximate label uncertainty by the intersection over union between a bounding box and its convex hull of the aggregated LiDAR observations.

Understanding how label errors distribute in a dataset is crucial in building reliable object detectors. Think about a car which is running directly in front of the ego-vehicle, and only has LiDAR observations on its back surface. It is difficult to determine object length and its longitudinal position, and we would intuitively downgrade the importance of such label both in training and testing. In [20], we make the first step to model the labelling errors for LiDAR-based object detection in self-driving datasets. We explicitly build a generative model to infer the distribution of 3D bounding box labels given the LiDAR points. The inferred label uncertainty not only reflects complex environmental noises inherent in LiDAR perception, such as typical L-shape and occlusion, but also shows the quality of bounding box labels in datasets.

**Contribution:** In this work, we combine the benefits of the KLD-loss [17] and the proposed label uncertainty [20] to improve a probabilistic LiDAR-based object detector which models uncertainty by the direct-modelling approach. By regularizing the predictive variances with our label uncertainty in the KLD-loss, we increase the detection accuracy in terms of Average Precision by up to 3.6% in the KITTI dataset [8], outperforming the methods with simple heuristics [12, 17]. Besides, we conduct an ablation study to show how label uncertainty affects training performance.

## 2  Methodology

In the following, we briefly introduce the probabilistic object detector in Sec. 2.1, discuss the KLD-loss to train the network in Sec. 2.2, and illustrate our method to estimate label uncertainty in Sec. 2.3. The method is shown in Fig. 1.

### 2.1  Probabilistic Object Detector

We employ probabilistic modelling with ProbPIXOR from [5]. It models the data-dependent uncertainty from PIXOR [22], a deterministic single stage object detection network using LiDAR point clouds (cf. Fig. 1). The network encodes the bounding box by its centroid positional offsets on the horizontal plane $\Delta x$, $\Delta y$, length and width in the log scale $\log(l), \log(w)$, and orientation $\sin(\theta), \cos(\theta)$. Given an input data point $\mathbf{x}$, ProbPIXOR assumes that each parameter of its ground truth bounding box $y^* \in [\Delta x, \Delta y, \log(l), \log(w), \sin(\theta), \cos(\theta)]$ follows a univariate Gaussian distribution, i.e. $q(y^*|\mathbf{x}) = \mathcal{N}(\hat{y}, \hat{\sigma}^2)$, with its mean and variance directly predicted by the network output layers. The network is trained by minimizing the negative log likelihood (NLL) of $q(y^*|\mathbf{x})$, resulting in an attenuated regression loss proposed by [13]:

$$\mathcal{L}_{\text{NLL}} = \frac{\log \hat{\sigma}^2}{2} + \frac{(y^* - \hat{y})^2}{2\hat{\sigma}^2}. \qquad (1)$$

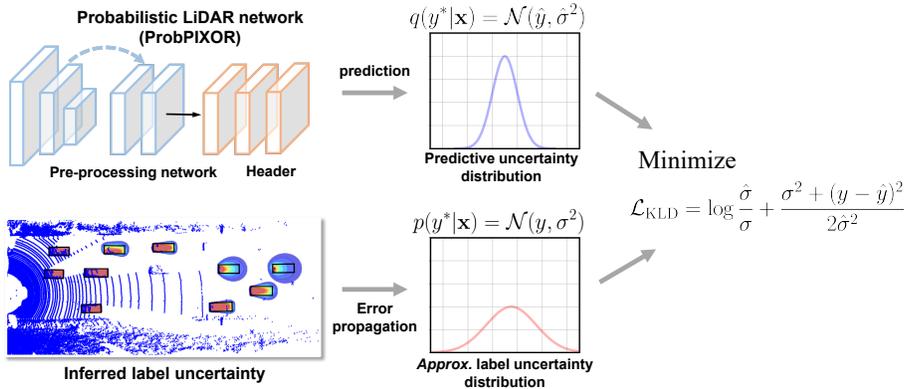

Fig. 1: An illustration of the proposed method. The probabilistic object detector ProbPIXOR [5] directly regresses the parameters of the network output probability $q(y^*|\mathbf{x})$, which is assumed to be Gaussian distributed. We also approximate the uncertainty inherent in ground truth labels [20] as the Gaussian distribution $p(y^*|\mathbf{x})$ via error propagation. The new loss function based on the Kullback-Leibler Divergence (KLD) $\text{KLD}(p,q)$ incorporates label uncertainty and regularizes predictive variances.

Note that the final regression loss is the summation of the attenuated regression losses for all bounding box variables.

### 2.2 Using the KLD-Loss to Incorporate Label Uncertainty

One of the major problems for Eq. 1 is that the variances $\hat{\sigma}^2$ are learnt in an un-supervised manner, without any ground truth information. This may lead to unstable training process or suboptimal detection performance, e.g. imagine that the network is predicting the low predictive variance for a training data point with error-prone label, or high predictive variance for an object from under-representative classes [21]. Recently, He *et al.* [12] and Meyer *et al.* [17] tackle this problem by minimizing the Kullback-Leiber Divergence (KLD-loss) between a prior probability distribution and the predictive probability distribution.

We follow the same idea and train ProbPIXOR with the KLD-loss. Let us treat $y^*$ mentioned above as an *unknown* ground truth parameter of a bounding box, and $y$ its (error-prone) label from human annotators. We assume that $y^*$ is corrupted from $y$ with a Gaussian noise characterized by its variance $\sigma^2$, i.e.

$p(y^*|\mathbf{x}) = \mathcal{N}(y, \sigma^2)$. To train ProbPIXOR, we minimize KLD between $p$ and $q$ with $\text{KLD}(p, q) = \int p(y^*|\mathbf{x}) \log(\frac{p(y^*|\mathbf{x})}{q(y^*|\mathbf{x})}) \mathrm{d}y^*$, resulting in a closed-form loss function [7]:

$$\mathcal{L}_{\text{KLD}} = \log \frac{\hat{\sigma}}{\sigma} + \frac{\sigma^2}{2\hat{\sigma}^2} + \frac{(y - \hat{y})^2}{2\hat{\sigma}^2}. \qquad (2)$$

The effectiveness of the KLD loss is largely dependent on the choice of label uncertainty $\sigma^2$. On the one hand, small $\sigma^2$ indicates that labels are accurate. In fact, when $\sigma^2 \to 0$, the label distribution becomes the Dirac-delta function, and the derivatives of Eq. 2 degenerate those of Eq. 1 [12]. On the other hand, large $\sigma^2$ indicates that labels are not trustful, and strongly regularize the predictive variances due to the $\frac{\sigma^2}{2\hat{\sigma}^2}$ term. Ideally, accurate samples are encouraged to train with low label uncertainty, and error-prone samples are penalized with large label uncertainty.

### 2.3 Inferring Label Uncertainty

In this work, we employ our previous method [20] to estimate label uncertainty, which is the posterior distribution $p(\mathbf{y}|\mathbf{x})$ of bounding box parameter vector $\mathbf{y}$ given all its $M$ associated LiDAR observations $\mathbf{x} = [x_1, x_2, x_i, ..., x_M]$. We assume that the mean of ground truth bounding box parameters $\mathbf{y}$ is accurate, and each LiDAR observation $x_i$ is generated by its $N$ nearest points uniformly sampled on the bounding box surface following a Gaussian Mixture Model (GMM) denoted as $p(x_i|\mathbf{y})$. Based on this, we approximate $p(\mathbf{y}|\mathbf{x})$ using Variational Bayes, resulting in a multi-variate Gaussian distribution (more details cf. [20]). The bottom left of Fig. 1 illustrates some label distributions in a scene, which reflects complex environmental noises such as L-shape and occlusion.

The full covariance matrix in the posterior distribution provides us correlations between two regression variables. For simplicity, however, we only consider their variances in this work, and ignore the correlations. Furthermore, the label encodings in the posterior distribution may not coincide with the bounding box encodings in the network. In this case, we use the error propagation technique [14] to transform uncertainty. For example, denote $c$ as a regression variable in a bounding box prediction, which can be expressed as a function of two variables $a$ and $b$ in the label uncertainty encoding, i.e. $c = f(a, b)$. The variance of $c$ is approximated by:

$$\sigma_c^2 = \left(\frac{\partial f}{\partial a}\right)^2 \sigma_a^2 + \left(\frac{\partial f}{\partial b}\right)^2 \sigma_b^2. \qquad (3)$$

Intuitively, we can also approximate label uncertainty with "num points" and "covx hull" heuristics. The method "num points" scales the label uncertainty according to the number of LiDAR observations within a ground truth bounding box, with the assumption that objects with increasingly sparse LiDAR points are more difficult to be labelled [3]. The method "covx hull" is proposed by [17], which approximates the label uncertainty by measuring the IoU value between a bounding box label and its convex hull of aggregated LiDAR points. In other words, this heuristic assumes that highly occluded objects tend to be error-prone. Our label uncertainty considers both the density and the shape of LiDAR point clouds, as illustrated in Fig. 2, and is more suitable to describe labelling noises.

---

[3] In practice, the number of points for an object vary significantly with a factor of $10^3$. Therefore, we calculate their logarithm values and do rescaling.

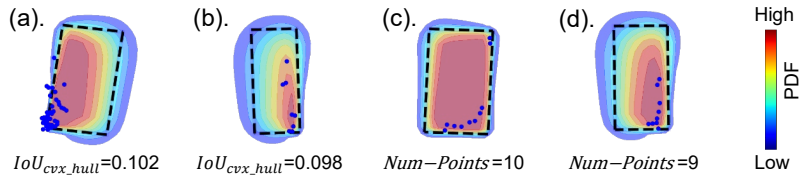

Fig. 2: Four labels with their diverse inferred distributions (shown by the heatmaps), but similar convex hull values (a-b) or similar number of LiDAR points (c-d).

## 3 Experimental Results

### 3.1 Experimental Setting

We validate our method on detecting "Car" objects in the KITTI dataset [8]. Following [1], we split the training data of 7481 frames into a *train* set and a *val* set, with approximately 50/50 ratio. All networks are optimized on the *train* set and evaluated on the *val* set. We use the LiDAR point cloud within the range length $\times$ width $\times$ height $= [0, 70]\text{m} \times [-40, 40]\text{m} \times [0, 2.5]\text{m}$, and do discretization with $0.1$ m resolution. We train networks with the SGD optimizer and the learning rate of $10^{-3}$ up to $140{,}000$ steps. Similar to PIXOR [22], we use global data augmentation by rotating and translating point clouds.

### 3.2 Detection Performance

We evaluate the detection performance with the average precision in the bird's eye view ($AP_{BEV}$) at the 0.7 Intersection over Union (IoU) threshold [8] and 41 recall steps. Results are illustrated in Tab. 1.

First, we build a baseline probabilistic object detector called "ProbPIXOR + $\mathcal{L}_{\text{NLL}}$" [5], which learns to predict uncertainty by minimizing the negative log likelihood (NLL)Eq. 1. It reaches better or on par detection performance compared to the original PIXOR network [22]. Note that "ProbPIXOR + $\mathcal{L}_{\text{NLL}}$" also corresponds to the method proposed by [12], which assumes a Dirac-delta function for label uncertainty.

Next, we train ProbPIXOR by the KLD-loss following Eq. 2, and compare among different methods differentiating from how they define label uncertainty (characterized by the variance $\sigma^2$). We compare among the label uncertainty with the fixed variance $\sigma^2 = 0.01$, the uncertainty extracted based on two simple heuristics "num points" and "covx hull" as discussed in Sec. 2.3, and the uncertainty from the proposed generative model ("Ours"). From the Table we observe that all networks based on the KLD-loss improve the detection performance compared to the baseline model, showing the benefits of the uncertainty regularization effect in the KLD-loss. Furthermore, the performance gains in the "Easy" and "Moderate" settings are larger than in the "Hard" setting for all KLD-based methods, because hard examples are usually associated with high label uncertainty and are penalized. More importantly, our method ("Ours") achieves the largest performance gain especially in the "Easy" setting by 3.6% AP compared to the baseline model. This indicates that the proposed generative model produces better label uncertainty than the simple heuristics that rely on the number of points or the convex hull.

Table 1: A comparison of detection performance in terms of $\text{AP}_{BEV}(\%)$. Highest performance gains compared to the baseline model are marked in bold.

| Methods | $\text{AP}_{BEV}(\%) \uparrow$ | | |
|---|---|---|---|
| | Easy | Moderate | Hard |
| PIXOR [22] | 86.79 | 80.75 | 76.60 |
| ProbPIXOR + $\mathcal{L}_{\text{NLL}}$ (Baseline) | 88.60 | 80.44 | 78.74 |
| ProbPIXOR + $\mathcal{L}_{\text{KLD}}$ ($\sigma^2 = 0.01$) | 89.09 (+0.49) | 81.33 (+0.89) | 79.17 (+0.43) |
| ProbPIXOR + $\mathcal{L}_{\text{KLD}}$ (num points) | 90.74 (+2.14) | 81.69 (+1.25) | 79.35 (+**0.61**) |
| ProbPIXOR + $\mathcal{L}_{\text{KLD}}$ (covx hull) | 90.05 (+1.45) | 81.12 (+0.68) | 78.84 (+0.10) |
| ProbPIXOR + $\mathcal{L}_{\text{KLD}}$ (**Ours**) | 92.22 (+**3.62**) | 82.03 (+**1.59**) | 79.16 (+0.42) |

### 3.3 Ablation Study

**The Choice of Label Uncertainty** As discussed in Sec. 2.2, the choice of label uncertainty has a big impact on the training procedure. In Fig. 3, we study how networks perform when being trained with increasing label uncertainty fixed for all training samples. The $AP_{BEV}$ scores remain similar at small label uncertainty ($\log(\sigma^2) < -2$), but drop significantly when $\log(\sigma^2) > -2$. This result suggests the necessity of setting a proper range of label uncertainty. Under-confident labels decrease detection accuracy.

**An Alternative of Using Label Uncertainty** A crucial component for training deep object detection models is data augmentation [2, 9], which can be categorized into global augmentation (e.g. rotating all LiDAR points in a frame) and local augmentation (e.g. randomly rescaling a ground truth). Similarly, we can locally augment ground truths by sampling according to their label uncertainties. In this way, training the network by maximizing the likelihood approximates to minimizing the KLD between the data distribution $p$ and the prediction distribution $q$ (cf. Appendix for detailed explanation). Tab. 2 shows the detection performance with this training strategy. We observe the (slightly) improved $AP$ scores compared to the baseline, indicating the benefit of data sampling. However, it still under-performs the methods based on the KLD-loss.

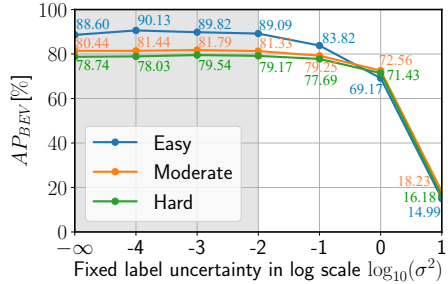

Fig. 3: The detection performance in $AP_{BEV}$ with increasing (fixed) label uncertainty characterized by its variance $\sigma^2$ in the log scale. All regression variables share the same variance for simplicity. Note that when $\sigma^2 \to 0$, i.e. $\log(\sigma^2) \to -\infty$, the derivatives of the KLD-loss degenerates to the standard NLL loss. The detection starts to drop by setting $\log(\sigma^2) > -2$.

Table 2: $AP_{BEV}(\%)$ for ProbPIXOR + $\mathcal{L}_{\text{NLL}}$ trained by sampling ground truths according to their label uncertainty. Relative improvements are compared to the baseline method.

| Easy | Moderate | Hard |
|------|----------|------|
| 89.39 (+0.79) | 81.21 (+0.77) | 78.89 (+0.15) |

## 4 Conclusion

This work leverages the previously proposed method [20] for estimating uncertainty inherent in ground truth bounding box parameters (which we call label uncertainty) to regularize the training of a probabilistic LiDAR-based object detector and to improve its detection accuracy. The method incorporates the inferred label uncertainty into the loss function that minimizes the Kullback-Leibler Divergence between predictive probability and label uncertainty. Experimental results on the KITTI dataset show the superiority of our method over the baseline model (without uncertainty regularization) or other models based on simple heuristics to estimate label uncertainty. In the future, we will verify our method in other state-of-the-art LiDAR-based object detectors, such as SA-SSD [11] and PV-RCNN [19]. We will also extend our method for multi-class detection, such as pedestrian and cyclist classes.

## Appendix

In this section, we show that training a network by maximizing the likelihood with ground truths sampled from their label uncertainties $p$ corresponds to minimizing the KLD between the label distribution $p$ and the prediction distribution $q$, as discussed in Sec. 3.3. More formally, we train the network by: Sample $y^* \sim p(y^*|\mathbf{x})$, $\max_q \big(\log\big(q(y^*|\mathbf{x})\big)\big)$, which is an approximation to $\max_q \mathbb{E}_{p(y^*|\mathbf{x})}\big[\log\big(q(y^*|\mathbf{x})\big)\big]$. We further have:

$$\max_q \mathbb{E}_{p(y^*|\mathbf{x})}\Big[\log\big(q(y^*|\mathbf{x})\big)\Big]$$

$$= \max_q \int p(y^*|\mathbf{x}) \log\big(q(y^*|\mathbf{x})\big) \mathrm{d}y^*$$

$$= \max_q \int p(y^*|\mathbf{x}) \log\Big(\frac{q(y^*|\mathbf{x})}{p(y^*|\mathbf{x})} p(y^*|\mathbf{x})\Big) \mathrm{d}y^*$$

$$= \max_q \int -p(y^*|\mathbf{x}) \log\Big(\frac{p(y^*|\mathbf{x})}{q(y^*|\mathbf{x})} \frac{1}{p(y^*|\mathbf{x})}\Big) \mathrm{d}y^*$$

$$= \min_q \Big(\underbrace{\int p(y^*|\mathbf{x}) \log\Big(\frac{p(y^*|\mathbf{x})}{q(y^*|\mathbf{x})}\Big) \mathrm{d}y^*}_{\mathrm{KLD}(p,q)} - \underbrace{\int p(y^*|\mathbf{x}) \log\big(p(y^*|\mathbf{x})\big) \mathrm{d}y^*}_{\text{Constant w.r.t. } q}\Big)$$

$$= \min_q \mathrm{KLD}(p,q)$$